\documentclass[10pt,letterpaper]{article}
\date{}

\usepackage{url}
\usepackage[utf8]{inputenc}
\usepackage[english]{babel}  
\usepackage{csquotes} 
\usepackage{textcomp}
\usepackage{gensymb}
\usepackage{subfigure}
\usepackage{amsmath}
\usepackage[table]{xcolor}
\usepackage{tabularx}
\usepackage{siunitx}
\usepackage[colorinlistoftodos,prependcaption,textsize=tiny]{todonotes}
\usepackage{multirow}

\newcommand{\madh}[1]{}%

\newcommand{\rotbox}[1]{\rotatebox[origin=l]{90}{#1}}

\newcommand{\withinTBE}[2][l]{\begin{tabular}{#1}#2\end{tabular}}
\def\xshift{4.45em}
\def\yshift{1.8em}
\definecolor{PHY}{HTML}{0068B4} 
\definecolor{PPROP}{HTML}{137F1A}
\definecolor{FPROP}{HTML}{7F1212}
\definecolor{SENSOR}{HTML}{A5B217}
          
\tikzstyle{element} = [draw, fill=PHY, minimum height=\yshift, minimum width=\xshift, text width=0.9*\xshift, rounded corners=0.5em, align=center, text=white]
\tikzstyle{edge} = [draw, line width=0.1em, -latex]
\tikzstyle{every node}=[font=\footnotesize]

\makeatletter
\renewcommand*{\@fnsymbol}[1]{\ensuremath{\ifcase#1\or 1\or 2\or *\or \ddagger\or
   		\mathsection\or \mathparagraph\or \|\or **\or \dagger\dagger
   		\or \ddagger\ddagger \else\@ctrerr\fi}}
\makeatother

\title{\LARGE \bf
Towards Robot-Centric Conceptual Knowledge Acquisition 
}

\author{Georg J\"ager\thanks{Georg J\"ager, Madhura Thosar and Sebastian Zug are with Faculty of Computer Science, Otto-von-Guericke University Magdeburg, Germany,
		{\tt\small gjaeger@ovgu.de, thosar@iks.cs.ovgu.de, zug@ivs.cs.uni-magdeburg.de}}\ \footnotemark[3] \ , Christian A. Mueller\thanks{Christian A. Mueller and Andreas Birk are with the Robotics Group, Computer Science \& Electrical Engineering Department, Jacobs University Bremen, Germany, {\tt\small \{chr.mueller,a.birk\}@jacobs-university.de}}\ \footnotemark[3]\ , Madhura Thosar\footnotemark[1]\ \thanks{These authors contributed equally to this work and share first authorship.}, \\ Sebastian Zug\footnotemark[1] \ and Andreas Birk\footnotemark[2]%
}

\begin{document}

\maketitle
\thispagestyle{empty}
\pagestyle{empty}

\begin{abstract}
Robots require knowledge about objects in order to efficiently perform various household tasks involving objects. %
The existing knowledge bases for robots acquire symbolic knowledge about objects from manually-coded external common sense knowledge bases such as ConceptNet, WordNet etc.
The problem with such approaches is the discrepancy between human-centric symbolic knowledge and robot-centric object perception due to its limited perception capabilities. 
Ultimately, significant portion of knowledge in the knowledge base remains ungrounded into robot's perception. %
To overcome this discrepancy, we propose an approach to enable robots to generate robot-centric symbolic knowledge about objects from their own sensory data, thus, allowing them to assemble their own conceptual understanding of objects.
With this goal in mind, the presented paper elaborates on the work-in-progress of the proposed approach followed by the preliminary results.

\end{abstract}

\section{Motivation}

Baber \cite{Baber2003_1} postulated that a deliberation for tool selection in humans or animals alike is facilitated by conceptual knowledge about objects, especially, knowledge about their physical and functional properties and relationship between them. 
The conceptual knowledge about household objects is desired in a service robot too when performing various household tasks, %
for instance, selecting an appropriate tool for a given task, selecting a substitute for a missing tool required in some task or action selection for using objects.
Since the demand for such conceptual knowledge about objects has been increasing, the development of such knowledge bases has been undertaken by the researchers around the world. 
In \cite{Thosar2018}, we reviewed the following existing knowledge bases developed for service robotics: KNOWROB \cite{Tenorth}, MLN-KB \cite{Zhu2014}, NMKB \cite{Pineda2017}, OMICS \cite{Gupta2004}, OMRKF \cite{Suh2007}, ORO \cite{Lemaignan2010}, OUR-K \cite{Lim2011}, PEIS-KB \cite{Daoutis2009}, and RoboBrain \cite{Saxena2014};  with respect to various criteria corresponding to the categories \textit{knowledge acquisition}, \textit{knowledge representation}, and \textit{knowledge grounding}.

\begin{figure}[tb]
   \centering
\begin{tikzpicture}

\node[element,text width=2*\xshift, minimum height=1.2*\yshift] (COMSENSOR) at (0,0) {Sensory Data};
\node[element,text width=2*\xshift, minimum height=1.2*\yshift] (COMKNOWLEDGE) at (0, 2.25*\yshift) {Common Sense Knowledge about Objects};

\node[element,text width=2*\xshift, minimum height=1.2*\yshift] (RCSENSOR) at (2.5*\xshift, 0.0*\yshift) {Sensory Data \& \\Extraction Methods};
\node[element,text width=2*\xshift, minimum height=1.2*\yshift] (RCCON) at (2.5*\xshift, 2.25*\yshift) {Conceptual Knowledge about Objects};

\path[edge] (COMKNOWLEDGE) edge node[xshift=0.35*\xshift, text width=0.5*\xshift] {Symbol \\Grounding} (COMSENSOR);

\path[edge] (RCSENSOR) edge node[xshift=-0.5*\xshift, align=right] {Symbol \\Generation} (RCCON);

\end{tikzpicture}
   \caption{Symbol grounding approach comparison: the typical approach to grounding knowledge vs. proposed approach to grounding the knowledge}
  \label{fig:grounding_approach}
\end{figure}
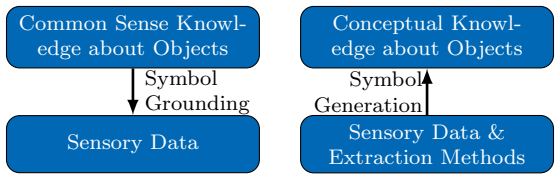

Knowledge grounding, a.k.a. symbol grounding or perceptual anchoring, which describes a mapping from abstract symbols to representative sensory data, is of special interest for robots. 
It closes the gap between symbolic reasoning, which enables abstract decision making, and interpreting their sensory data, which is imperative for understanding a robot's environment.
In many existing knowledge bases (see \cite{Thosar2018} for details), symbols are defined by humans and are derived from common sense knowledge bases such as WordNet (KnowRob, MLN-KB, OMICS, RoboBrain), %
Cyc (PEIS-KB) %
OpenCyc (KnowRob, ORO, RoboBrain) (see left side of Fig.~\ref{fig:grounding_approach}).
By employing sensors, a mapping from these symbols to sensory data is constructed. 
However, %
this imposes a human perspective on a robot's sensory data. 
Humans select sensors and dictate their interpretation for grounding the symbols. 
This can only work with unambiguous sensory data and complete knowledge about all relevant object categories.

In contrast, we propose a bottom-up approach to symbol grounding, a robot-centric symbol generation approach, see right side of Fig.~\ref{fig:grounding_approach}. 
Our current research work is aimed at creating a multi-layered dataset that can be used to build robot-centric conceptual knowledge about household objects. %
Instead of predefining the symbols and grounding them afterwards, we use sensory data and robot-centric extraction methods to generate qualitative data for each object property. 
Afterwards, symbols are generated from this qualitative data in an unsupervised manner and thereby inherently grounded in the robots sensory data. 
Moreover, this approach enables a robot to build up its own individualized understanding about objects that does not rely on completeness.

The conceptual knowledge considered in this work primarily involves properties of the objects.
The properties considered are divided into \emph{physical} and \emph{functional} properties where physical properties describe the physicality of the objects such as \emph{rigidity}, \emph{weight}, \emph{hollowness} while the functional properties ascribe the (functional) abilities or affordances to the objects such as \emph{containment}, \emph{blockage}, \emph{movability}.

The remainder of this paper is structured as follows:
we initially define the considered physical and functional properties and introduce the property acquisition in Sec.~\ref{sec:property_acquisition}.
Using the presented definitions and acquisition methods, we further elaborate on the architecture of our framework for the generation of robot-centric knowledge in Sec.~\ref{sec:framework}.
In Sec.~\ref{sec:preliminary} we present our preliminary results.
Finally, we conclude our work and discuss possible future work (Sec.~\ref{sec:future}).

\section{Property Acquisition}
\label{sec:property_acquisition}

As a prerequisite to the generation of a robot-centric knowledge base, we present definitions of the considered object properties in Sec.~\ref{subsec:property_definition}. 
These are general and not specialized towards a robotic platform as they represent the humans' perspective. 
In contrast, in Sec.~\ref{subsec:extraction_methods} we define methods for acquiring scalar representations of the defined object properties for a %
robotic platform. 
Their implementations are embedded into an extensible framework for data aggregation, which we will briefly introduce in Sec.~\ref{sec:framework}.

\subsection{Property Definition}
\label{subsec:property_definition}

Overall, we consider ten core object properties.
We start with six physical properties. 
These form the basis from which the remaining four functional properties, which we will define afterwards, emerge from.

\subsubsection{Physical Properties}
\label{subsec:physical_properties}

As a selection of core physical properties linked to the physicality of an object we have considered \textbf{\textit{size}}, \textit{\textbf{hollowness}}, \textit{\textbf{flatness}}, \textit{\textbf{rigidity}}, \textit{\textbf{roughness}}, and \textit{\textbf{weight}}.
This selection is inspired from the discussion on the design of tools offered by Baber in \cite{Baber2003_6} where it is stated that, among others, the properties such as \textit{\textbf{shape}}, \textbf{\textit{size}}, \textbf{\textit{rigidity}}, \textbf{\textit{roughness}}, and \textbf{\textit{weight}} play a significant role in the design of a tool.
In the following, we discuss briefly the property definitions which state how they are to be measured. 

\textit{\textbf{Size}} of an object is described by its spatial dimensionality in form of %
\emph{length}, \emph{width} and \emph{height}.
\textit{\textbf{Flatness}}, on the contrary, describes a particular aspect of an object's shape. 
\textbf{\textit{Flatness}} is defined as the ratio between the area of an object's greatest plane and its overall surface area. For instance, a sheet of paper has maximal \textbf{\textit{flatness}} while a ball has minimal \textbf{\textit{flatness}}.
\textit{\textbf{Hollowness}} focuses on another aspect of an object's shape. 
It is the amount of visible cavity or empty space within an object's enclosed volume. 
\textit{\textbf{Weight}} of an object 
is borrowed from physics: the object's \textbf{\textit{weight}} is the force acting on its mass within a gravitational field. 
Similar to gravity, forces acting on objects might deform it depending on its \textit{\textbf{rigidity}}. 
Consequently, we define \textit{\textbf{rigidity}} as the degree of deformation caused by a force operating vertically on an object. 
The last physical property to be defined is \textit{\textbf{roughness}}. It provides a feedback about the object's surface. 
Therefore, we simplify the physical idea of friction and define \textit{\textbf{roughness}} as an object's resistance to slide.

\subsubsection{Functional Properties}
\label{subsec:functional_properties}

Opposed to physical properties, functional properties describe the functional capabilities or affordances of objects.
It is proposed in \cite{Baber2003_5} that functional properties do not exist in isolation, rather certain physical properties are required to enable them.
The proposed system follows the same suit where each functional property is defined in terms of the associated physical properties

A basic functional property is \textit{\textbf{support}}. It describes an object's capability to \textbf{\textit{support}}, i.e. to carry another object. 
Therefore, an object is attributed with \textbf{\textit{support}}, if other objects can be stably placed on top of the supporting object.
The \textit{\textbf{containment}} property extends this idea. 
An object is attributed with property \textit{\textbf{containment}} if it can encompass another object to a certain degree.
Finally, we also consider \textit{\textbf{movability}}, which describes the required effort to move an object, and \textit{\textbf{blockage}}, which describes an object's capability to stop the movement of another object.

\subsection{Property Extraction}
\label{subsec:extraction_methods}

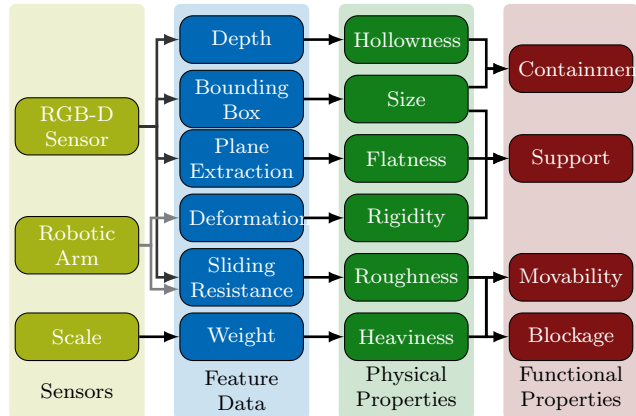
\begin{figure}[tb]
\centering
\def\xfactor{1.4}

\begin{tikzpicture}

\node[fill=SENSOR!20, minimum height=8.75*\yshift, minimum width=1.15*\xshift,rounded corners=0.25em,label={[yshift=-8.5*\yshift, text width=1.15*\xshift, align=center]Sensors}] (FEATURE) at (-1*\xfactor*\xshift, -3.625*\yshift) {};

\node[fill=PHY!20, minimum height=8.75*\yshift, minimum width=1.15*\xshift,rounded corners=0.25em,label={[yshift=-8.75*\yshift, text width=1.15*\xshift, align=center]Feature\\Data}] (FEATURE) at (0*\xfactor*\xshift, -3.625*\yshift) {};

\node[fill=PPROP!20, minimum height=8.75*\yshift, minimum width=1.15*\xshift,rounded corners=0.25em,label={[yshift=-8.75*\yshift, text width=1.15*\xshift, align=center]Physical\\ Properties}] (PHYPROP) at (1*\xfactor*\xshift, -3.625*\yshift) {};

\node[fill=FPROP!20, minimum height=8.75*\yshift, minimum width=1.15*\xshift,rounded corners=0.25em,label={[yshift=-8.75*\yshift, text width=1.15*\xshift, align=center]Functional \\Properties}] (FUNPROP) at (2*\xfactor*\xshift, -3.625*\yshift) {};

\node[element, fill=SENSOR] (SEN1) at (-1*\xfactor*\xshift,-1.825*\yshift) {RGB-D Sensor};
\node[element, fill=SENSOR] (SEN2) at (-1*\xfactor*\xshift,-4.325*\yshift) {Robotic Arm};
\node[element, fill=SENSOR] (SEN3) at (-1*\xfactor*\xshift,-6.25*\yshift) {Scale};

\node[element, fill=PHY] (PHY1) at (0,0) {Depth};
\node[element, fill=PHY] (PHY2) at (0,-1.25*\yshift) {Bounding Box};
\node[element, fill=PHY] (PHY3) at (0,-2.5*\yshift) {Plane Extraction};
\node[element, fill=PHY] (PHY4) at (0,-3.75*\yshift) {Deformation}; 
\node[element, fill=PHY] (PHY5) at (0,-5*\yshift) {Sliding Resistance};
\node[element, fill=PHY] (PHY6) at (0,-6.25*\yshift) {Weight};

\node[element, fill=PPROP] (PPROP1) at (1*\xfactor*\xshift,0) {Hollowness}; 
\node[element, fill=PPROP] (PPROP2) at (1*\xfactor*\xshift,-1.25*\yshift) {Size};
\node[element, fill=PPROP] (PPROP3) at (1*\xfactor*\xshift,-2.5*\yshift) {Flatness};
\node[element, fill=PPROP] (PPROP4) at (1*\xfactor*\xshift,-3.75*\yshift) {Rigidity};
\node[element, fill=PPROP] (PPROP5) at (1*\xfactor*\xshift,-5*\yshift) {Roughness};
\node[element, fill=PPROP] (PPROP6) at (1*\xfactor*\xshift,-6.25*\yshift) {Heaviness};

\node[element, fill=FPROP] (FPROP1) at (2*\xfactor*\xshift,-0.6125*\yshift) {Containment}; 
\node[element, fill=FPROP] (FPROP2) at (2*\xfactor*\xshift,-2.5*\yshift) {Support};
\node[element, fill=FPROP] (FPROP3) at (2*\xfactor*\xshift,-5*\yshift) {Movability};
\node[element, fill=FPROP] (FPROP4) at (2*\xfactor*\xshift,-6.25*\yshift) {Blockage};

\path[edge,draw=black!80] (SEN1.east) -- ++(0.15*\xshift,0)|- (PHY1.west);
\path[edge,draw=black!80] (SEN1.east) -- ++(0.15*\xshift,0)|- (PHY2.west);
\path[edge,draw=black!80] (SEN1.east) -- ++(0.15*\xshift,0)|- (PHY3.west);
\path[edge,draw=black!80] (SEN1.east) -- ++(0.15*\xshift,0)|- (PHY5.west);

\path[edge,draw=black!50] (SEN2.east) -- ++(0.1*\xshift,0)|- (PHY4.west);
\path[edge,draw=black!50] (SEN2.east) -- ++(0.1*\xshift,0)|- ([yshift=-0.25*\yshift]PHY5.west);

\path[edge,draw=black] (SEN3.east) -- ++(0.15*\xshift,0)|- (PHY6.west);

\path[edge] (PHY1) -> (PPROP1);
\path[edge] (PHY2) -> (PPROP2);
\path[edge] (PHY3) -> (PPROP3);
\path[edge] (PHY4) -> (PPROP4);
\path[edge] (PHY5) -> (PPROP5);
\path[edge] (PHY6) -> (PPROP6);

\path[edge] (PPROP1.east) -- ++(0.15*\xshift,0)|- (FPROP1.west);
\path[edge] ([yshift=0.25*\yshift]PPROP2.east) -- ++(0.15*\xshift,0)|- (FPROP1.west);

\path[edge] ([yshift=-0.25*\yshift]PPROP2.east) -- ++(0.15*\xshift,0)|- (FPROP2.west);
\path[edge] (PPROP3.east) -- ++(0.15*\xshift,0)|- (FPROP2.west);
\path[edge] (PPROP4.east) -- ++(0.15*\xshift,0)|- (FPROP2.west);

\path[edge] (PPROP5.east) -- ++(0.15*\xshift,0)|- (FPROP3.west);
\path[edge] (PPROP5.east) -- ++(0.15*\xshift,0)|- (FPROP4.west);
\path[edge] (PPROP6.east) -- ++(0.15*\xshift,0)|- (FPROP3.west);

\end{tikzpicture}
\caption{Proposed property hierarchy and their dependencies.}
\label{fig:property_hierarchy} 
\end{figure}

\begin{figure}[tb]
	\centering
    \subfigure[Multi-camera setup]{\label{fig:multicamsetup}\includegraphics[height=4cm]{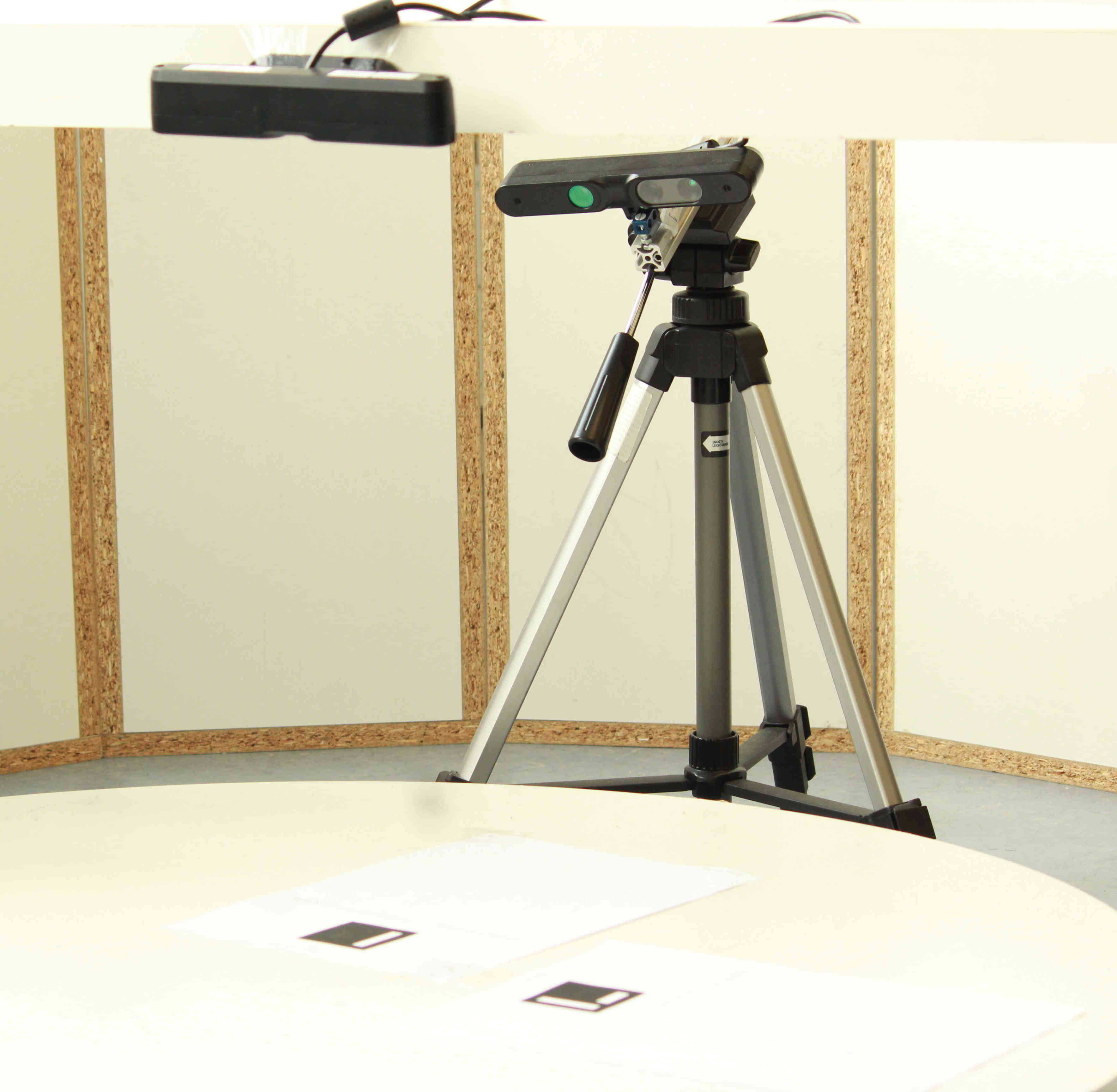}} 
\subfigure[Side RGB-D camera]{\label{fig:sample_box_side_cam}\includegraphics[height=4cm]{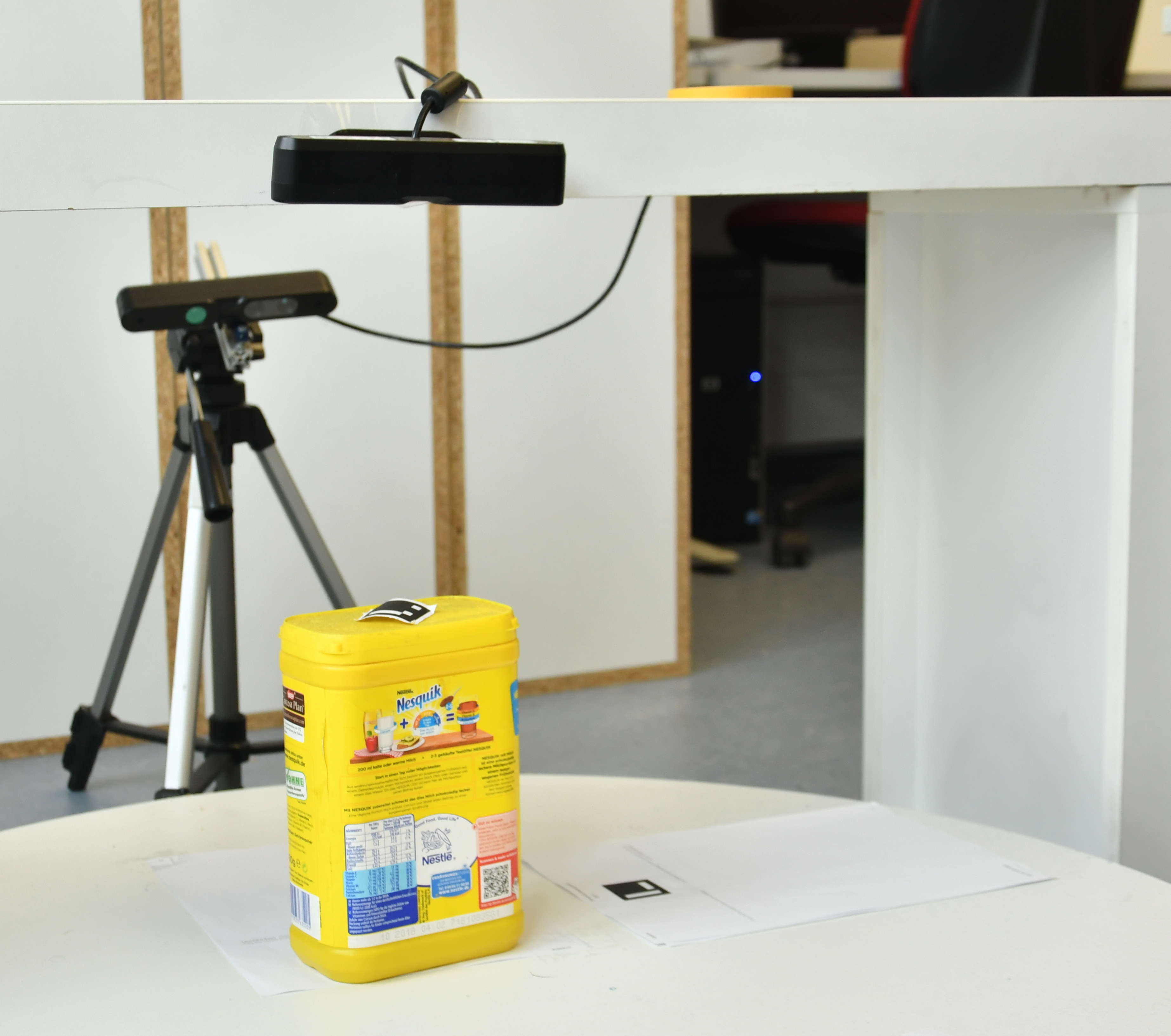}}	
    
    \subfigure[Top RGB-D camera]{\label{fig:sample_bowl_top_cam}\includegraphics[height=4cm]{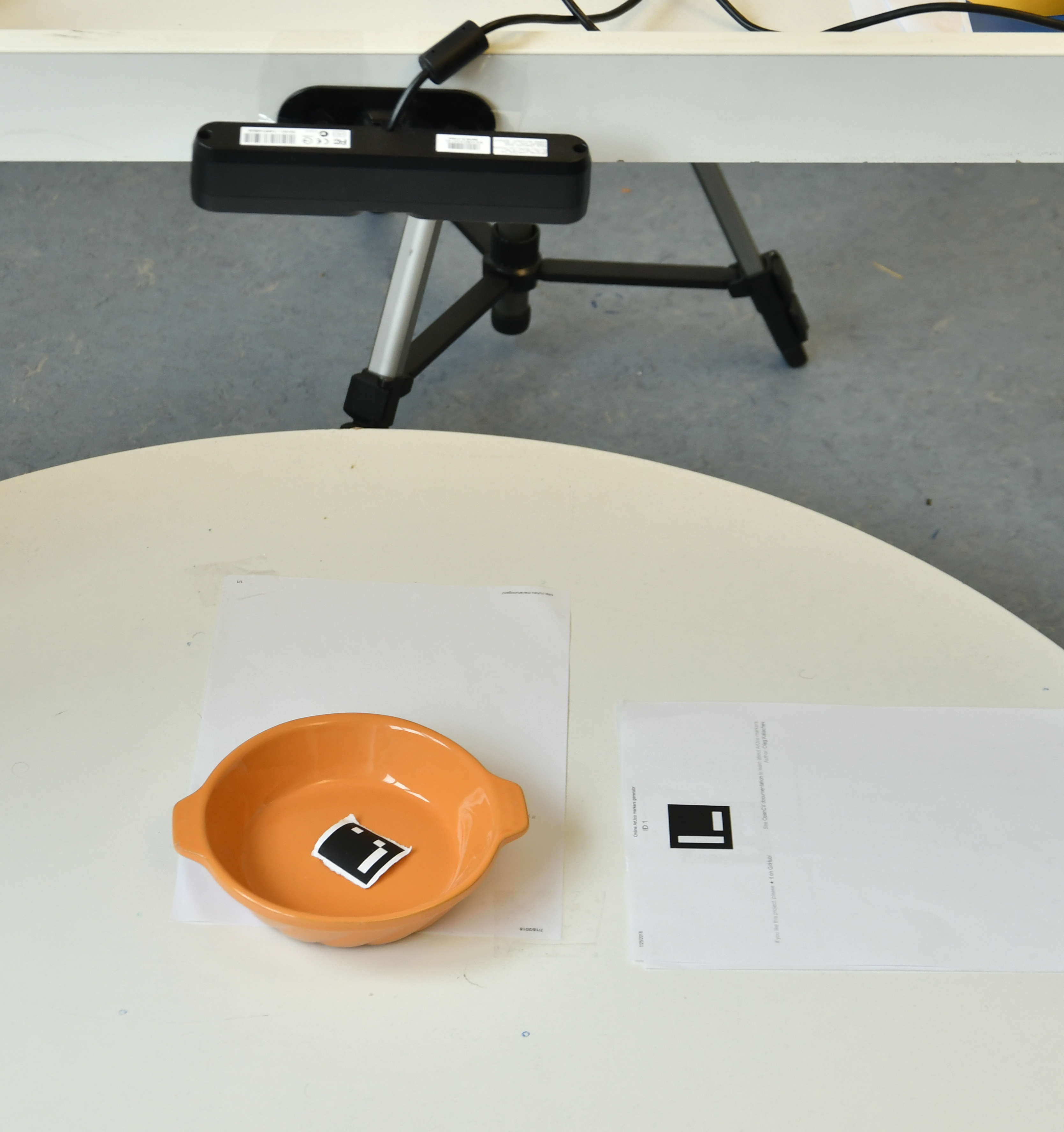}}
    	\subfigure[Top RGB-D camera]{\label{fig:sample_bowl_top_cam2}\includegraphics[height=4cm]{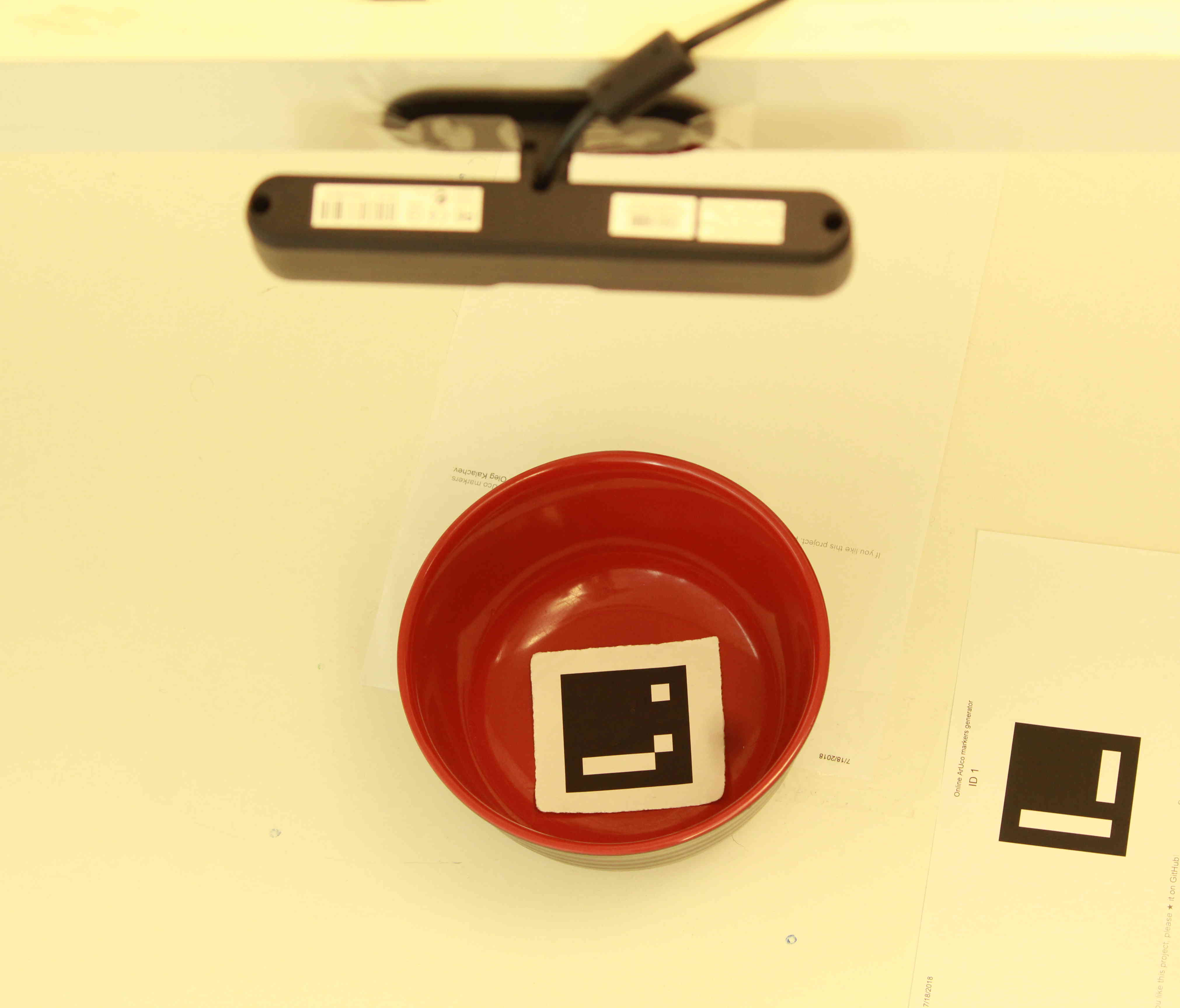}}
	\caption{Experimental setup consisting of a two camera combination, for acquiring physical properties such as \textit{\textbf{hollowness}}, \textit{\textbf{size}}, \textit{\textbf{roughness}}.}
	\label{fig:setup:physical_property}
\end{figure}

Though the property definitions in Sec.~\ref{subsec:property_definition} are %
formulated from a human perspective, 
we aim at enabling a robot to assemble its own understanding about objects. 
Hence, we have devised the extraction methods allowing a robot to interpret its sensory data for generating scalar representations of object properties. %
The different levels of abstractions, starting with the sensory data and ending with functional properties, are shown in Fig.~\ref{fig:property_hierarchy}.
This, however, requires to take the available sensors and actuators into account to ensure observability of all properties. 
While this means that the presented methods are tuned towards our robotic platform (a Kuka youBot~\cite{Bischoff2011} (see Fig~\ref{fig:exp:roughness5}) and a Asus Xtion Pro depth sensor~\cite{Swoboda2014} (see Fig.~\ref{fig:multicamsetup})), they are adoptable to other robotic platforms as we use common hardware.

Across all methods, we assume that the object for which the property shall be measured, is placed in its most natural position. 
For instance, a cup is most commonly placed in such a way, that its opening points upwards.

\subsubsection{Physical Properties}
The \textit{\textbf{size}} of an object is extracted from point clouds of an RGB-D sensor by segmenting an object from the scene which then is used to estimate the bounding box.
The length, width, and height of the bounding box are then used to measure the \textit{\textbf{size}}.

\begin{figure}[tb]
	\centering	\subfigure[Rigidity extraction at time $t\mathrm{=}1$]{\label{fig:exp:rigidity1}\includegraphics[height=2.4cm]{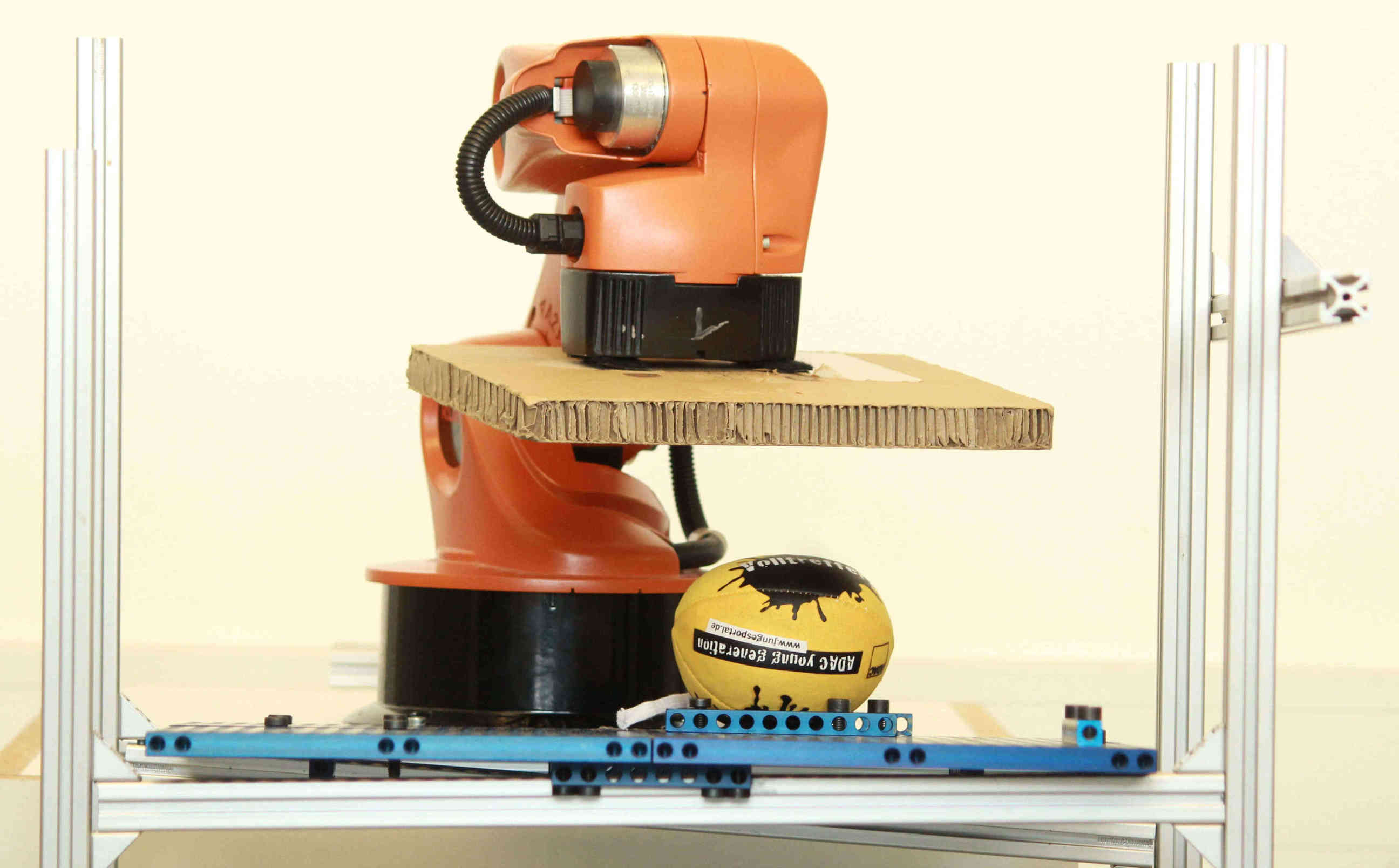}}
	\subfigure[Rigidity extraction at time $t\mathrm{=}2$]{\label{fig:exp:rigidity3}\includegraphics[height=2.4cm]{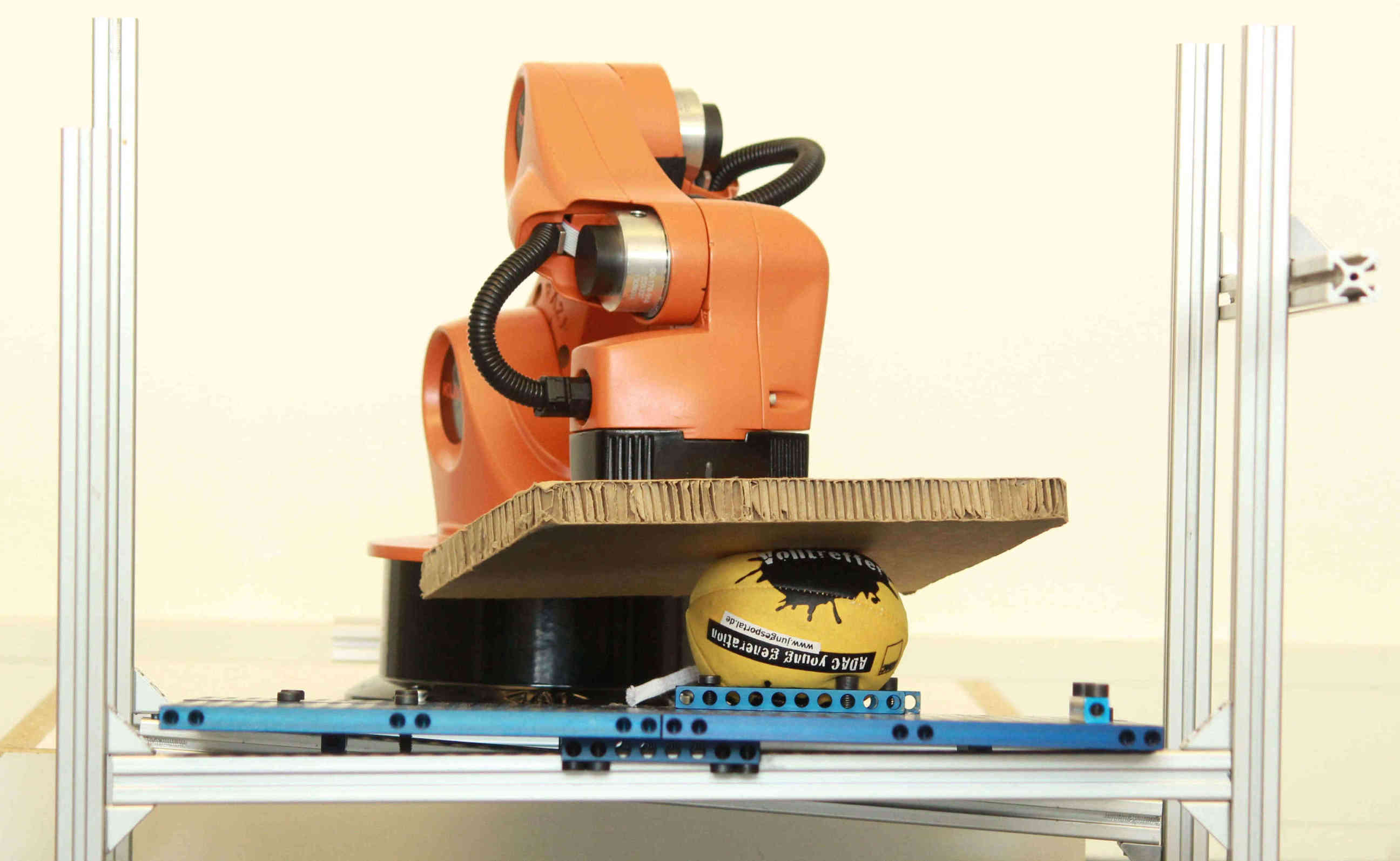}}
    \subfigure[Rigidity extraction at time $t\mathrm{=}3$]{\label{fig:exp:rigidity2}\includegraphics[height=2.4cm]{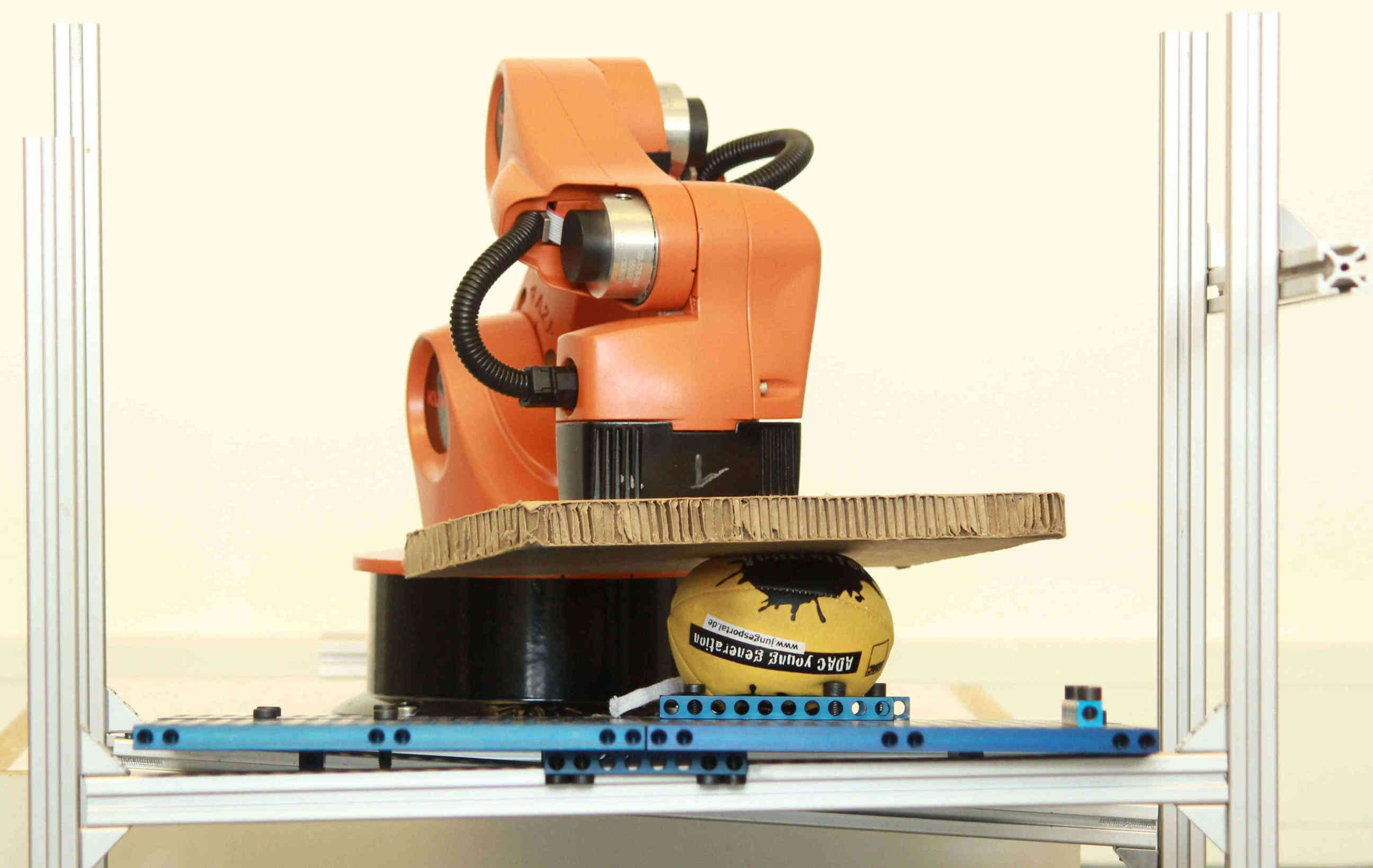}}
    
\subfigure[Roughness extraction at time $t\mathrm{=}1$]{\label{fig:exp:roughness5}\includegraphics[height=2.575cm]{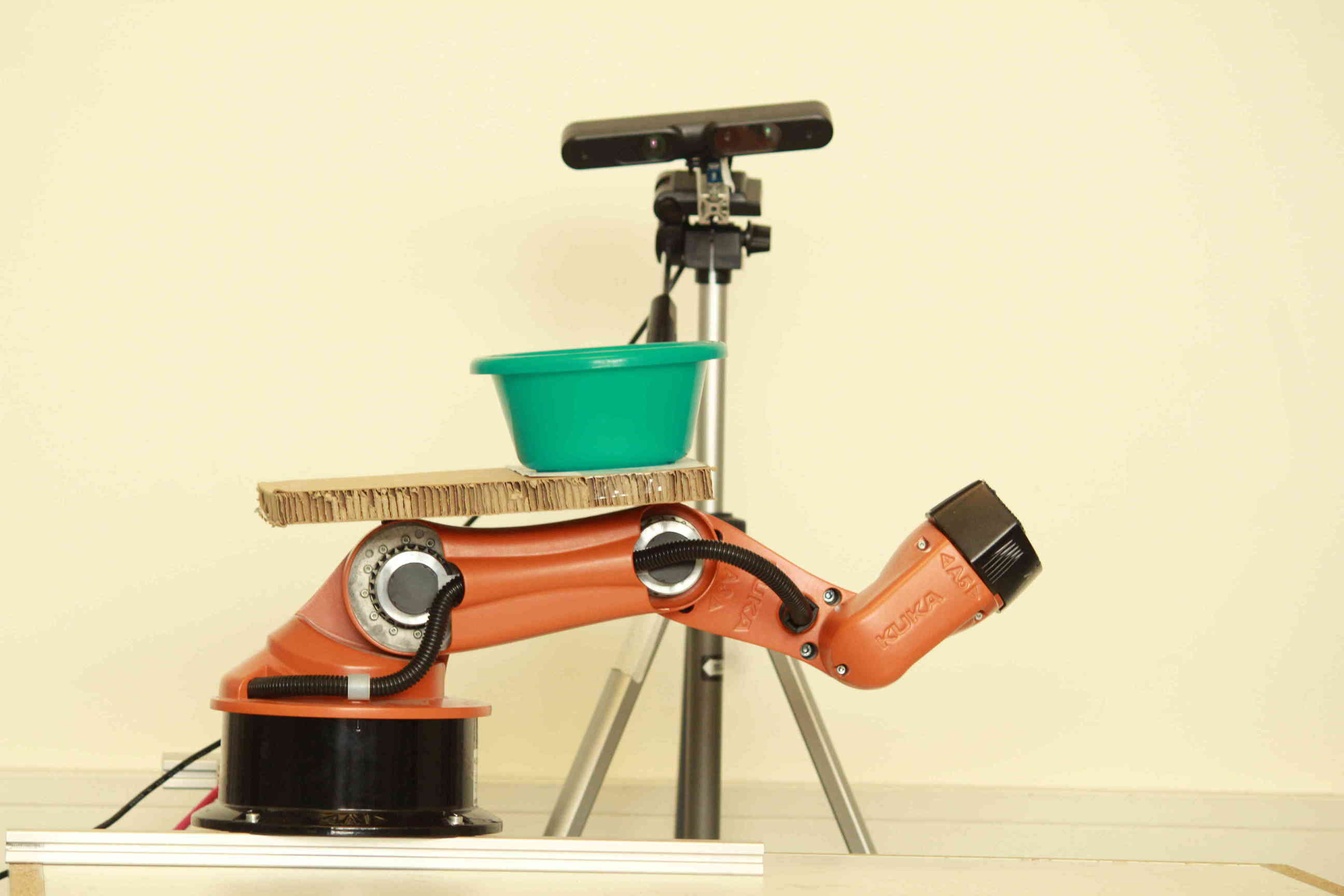}}
	\subfigure[Roughness extraction at time $t\mathrm{=}2$]{\label{fig:exp:roughness6}\includegraphics[height=2.575cm]{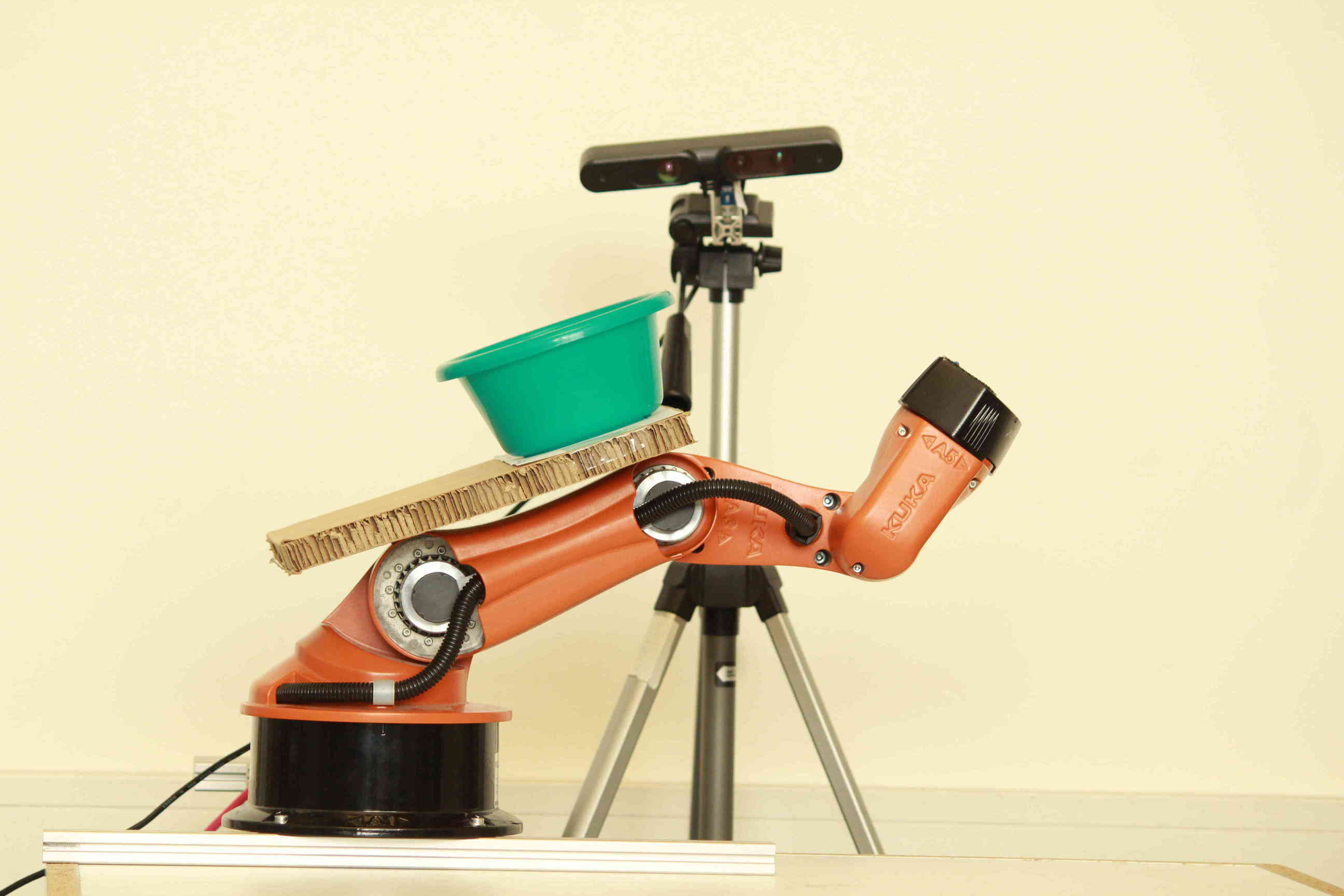}}
\subfigure[Roughness extract. at time $t\mathrm{=}3$]{\label{fig:exp:roughness7}\includegraphics[height=2.575cm]{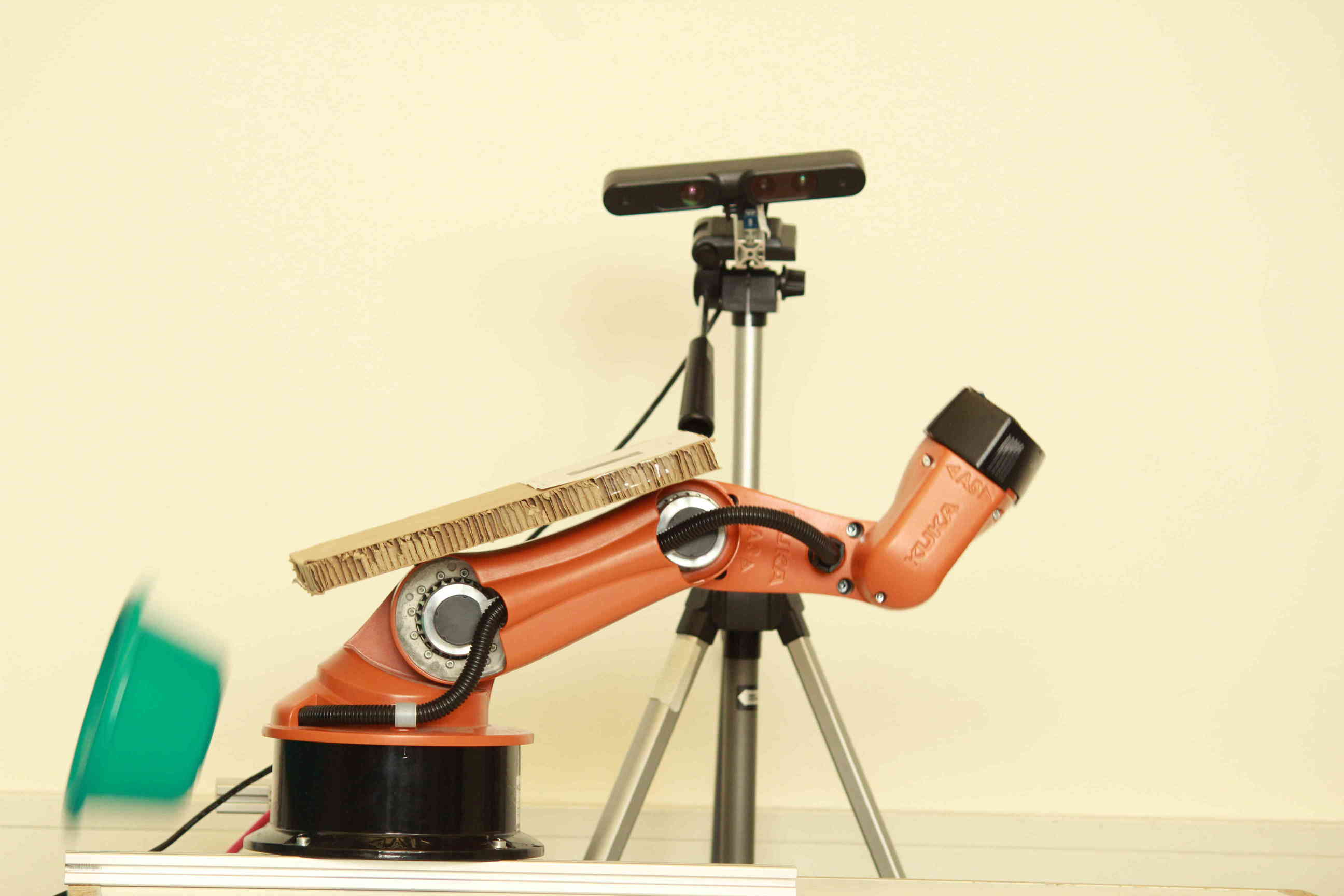}}
\caption{Experimental setup consisting of a camera-manipulator combination, for acquiring physical properties such as \textit{\textbf{rigidity}} and \textit{\textbf{roughness}} which are required to acquire functional properties such as \textit{\textbf{support}} and \textit{\textbf{movability}}.}
	\label{fig:setup:physical_property_cam_robot}
\end{figure}

Additionally, the segmented object point cloud enables the robot to extract the object's \textit{\textbf{flatness}} value.
By observing an object from above, an object's greatest top-level plane is extracted using RANSAC (RAndom SAmple Consensus). 
The size of this plane, that is, the number of points corresponding to this plane is divided by the number of points representing the observed object to get a scalar measure of its \textbf{\textit{flatness}}.

Similar to \textit{\textbf{flatness}}, \textit{\textbf{hollowness}} is also focused on an object's shape. According to its definition, an object's enclosed, but not filled volume defines its hollowness. %
For the sake of simplicity, we measure the internal depth and height of an object, which resembles the enclosed volume, and use their ratio as \textbf{\textit{hollowness}} value. 
To measure depth and height, we employ marker detection, which is robust and insusceptible to noise. 
We place one marker inside (or on top) of the considered object and another marker right next to the object which functions as a global spatial reference (see samples in Fig.~\ref{fig:sample_box_side_cam}-\subref{fig:sample_bowl_top_cam2}).%

In contrast, measuring the \textit{\textbf{weight}} of an object is straight forward. 
Using a scale with a resolution of 1\si{\g}, the actual \textbf{\textit{weight}} of each object can be measured directly. While this requires additional hardware, a robot could easily try to lift an object and calculate its \textbf{\textit{weight}} by converting the efforts observed during the process in each of its joints (we are currently working on replacing the method). 

The \textit{\textbf{rigidity}} property requires a more sophisticated measuring process. 
Following its definition, we use a robotic arm with a planar end-effector to vertically exert a force onto an object until predefined efforts in the arm's joints are exceeded, see Fig.~\ref{fig:exp:rigidity1}-\subref{fig:exp:rigidity2}.
Using the joints' positions at the first contact with the object and when the efforts are exceeded, the vertical movement of the arm during the experiment is calculated. For rigid objects, this distance is zero while it is increased continuously for non-rigid objects.

Similar to \textit{\textbf{rigidity}}, \textit{\textbf{roughness}} requires interaction to measure an object's resistance to sliding. 
Therefore, the robotic arm is used to act as a ramp on which the considered object is placed, see Fig.~\ref{fig:exp:roughness5}-\subref{fig:exp:roughness7}. Starting horizontally, with an angle of $0\degree$, the ramp's angle is increased and thereby causes the force pulling the object down the ramp to be increased too. As soon as the object starts sliding, which is detected based on marker detection, the movement is stopped. In this position, the ramp's angle is a measure of the object's \textbf{\textit{roughness}}.

\subsubsection{Functional Properties}
As functional properties are enabled by an object's physical attributes, we define their extraction methods on the basis of an object's physical properties. %
Corresponding to its definition, \textit{\textbf{support}} requires to consider three aspects of an object. Firstly, the considered object needs to be rigid. Secondly, for carrying another object, the sizes of both need to match. Thirdly, the object's shape needs to be sufficiently flat in order to enable the placing of another object on top of it. Consequently, for measuring support, we consider \textit{\textbf{rigidity}}, \textit{\textbf{size}}, and \textit{\textbf{flatness}}.

Similarly, the \textit{\textbf{containment}} property requires to consider two aspects. In order to contain something, an object needs to be hollow. On the other hand, it's \textbf{\textit{size}} itself needs to be respected when considering whether it can contain another object. Thus, the value of an object's \textit{\textbf{containment}} property is formed by combining its \textit{\textbf{hollowness}} and \textit{\textbf{size}}.

Extracting an object's \textit{\textbf{movability}} is based on a robot's primary ways of moving objects: either by lifting or pushing.
For both, an object's \textit{\textbf{weight}} is important. Additionally, when pushing an object, its sliding resistance, that is, its \textit{\textbf{roughness}} (see Fig.~\ref{fig:setup:physical_property_cam_robot}), needs to be considered.

Finally, assessing an object's \textit{\textbf{blockage}}, can be derived from its \textit{\textbf{movability}}.
According to its definition, \textbf{\textit{blockage}} states to which degree an object is able to stop another object's movement. Thus, the object itself needs to be not movable by the other object, which is the inverse of its \textit{\textbf{movability}}.

The described hierarchy of object properties as well as their dependency on feature and sensory data is illustrated in Fig.~\ref{fig:property_hierarchy}.

\subsection{Framework for Dataset Acquisition}
\label{sec:framework}

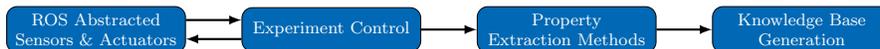
\begin{figure}[h!]
\centering
\scalebox{0.8}{
\begin{tikzpicture}

\node[element,text width=1.75*\xshift] (ROS) at (0,0) {ROS Abstracted Sensors \& Actuators};
\node[element,text width=1.75*\xshift, minimum height=1.15*\yshift] (DATA) at (2.5*\xshift, 0) {Experiment Control};
\node[element,text width=1.75*\xshift] (EXTRACT) at (5*\xshift,0) {Property \\Extraction Methods};
\node[element,text width=1.75*\xshift] (KNOWLEDGE) at (7.5*\xshift,0) {Knowledge Base Generation};

\path[edge] (ROS) edge[transform canvas={yshift=0.25*\yshift}]  (DATA);
\path[edge] (DATA) edge[transform canvas={yshift=-0.25*\yshift}]  (ROS);
\path[edge] (DATA) edge (EXTRACT);
\path[edge] (EXTRACT) edge (KNOWLEDGE);

\end{tikzpicture}
}
\caption{Data flow within the data set creation framework.}
\label{fig:framework}
\end{figure}

Using the extraction methods of Sec.~\ref{subsec:extraction_methods}, we present our ROS-based framework with which our robotic platform gathers data about objects to build up its individualized knowledge. A schematic overview on the framework is given by Fig.~\ref{fig:framework}. %

\begin{table}%
	\caption{Human-predefined object classes, number of instances in each class and their numeric labels used in the plots in Fig.~\ref{fig:p_isomap} and~\ref{fig:s_isomap}}
	\label{tab:dataset_label_definitions}
	\centering
	\footnotesize
	\setlength{\tabcolsep}{0.2em}
	\begin{tabular}{|l|*{17}{c|} | c |}
		\hline
		\rotatebox[origin=l]{45}{Class} & 
		\rotbox{Plastic Box} &
        \rotbox{Paper Plate} & 
        \rotbox{Steel Cup} & 
        \rotbox{Ceramic Bowl} & 
        \rotbox{Plastic Cup}  & 
        \rotbox{Paper Box} & 
        \rotbox{Ceramic Plate} & 
        \rotbox{Ball} & 
        \rotbox{Metal Box} & 
        \rotbox{Paper Cup} &
        \rotbox{Marker}  &  
        \rotbox{Plastic Bowl} & 
        \rotbox{Ceramic Cup} & 
        \rotbox{Sponge} &
        \rotbox{Marble Plank} &
        \rotbox{Ceramic Glass} &
        \rotbox{Book} & \cellcolor{black!20} \withinTBE[c]{$\sum$\vspace*{1em}} \\ \hline

      \rotatebox[origin=c]{45}{Label} & 0 & 1 & 2 & 3 & 4 & 5 & 6 & 7 & 8 & 9 & 10 & 11 & 12 & 13 & 14 & 15 & 16 & \cellcolor{black!20} 17
	  \\ \hline	
      \rotatebox[origin=c]{45}{\#} & 1 & 1 & 1 & 3 & 3 & 9 & 6 &1 &3 & 1 &3 &3 &3 &2 &1 &1 &4& \cellcolor{black!20} 46	  \\ \hline	
	\end{tabular}
\end{table}

Although different software platforms for operating robots exist (e.g. Fawkes \cite{Niemueller2010} or Orocos~\cite{Soetens2006}), ROS (Robot Operating System)~\cite{Koubaa2017} became a quasi-standard. 
Given the amount of supported hardware components, building on top of this middleware enables other researchers to reproduce our results and adapt our framework according to their specific hardware (which is essential for building an robot-centric knowledge base). 

Consequently, the interface for operating sensors and actuators is provided to our framework by ROS. This interface is used by different experiments for observing and interacting with objects to acquire the necessary sensory data. Together both blocks (\textit{ROS Abstracted Sensors \& Actuators} and \textit{Experiment Control}) form a control loop which generates feature data (see Fig.~\ref{fig:property_hierarchy}).

To provide extensibility and comparability along with our framework, versatile experiments can be defined by either adding independent ROS nodes or extending existing experiments. The generated feature data is further processed by property extraction methods to calculate the values of each property for the object currently under consideration. Again, extendability and comparability are facilitated by running each extraction method as an independent ROS node and therefore by providing a \textit{plug-and-play} interface.

The data generated by the extraction methods resembles the scalar representations of an objects properties. Therefore, physical as well as functional properties of objects are available and used by the knowledge base generation process to generate a symbolic representation.

\section{Knowledge Base}
\label{sec:knowledge_generation}

Using the framework described in the previous section, we can employ our robotic platform to gather scalar data about an object's properties. 
However, this data can not be used for symbolic reasoning yet. 
To facilitate this application, a knowledge base needs to be generated.
We briefly describe this step in this section.

The knowledge base primarily consists of two layers: knowledge about object instances and knowledge about object classes.
The primary input to the knowledge base is the data about the physical properties of the objects where each object instance is represented in terms of its physical properties as well as its functional properties.
The data about the properties of the objects is processed by the knowledge base module in two stages: \textit{sub-categorization} and \textit{conceptualization}.
In the sub-categorization process, the non-symbolic continuous data of each property is transformed into symbolic data using a clustering algorithm such as K-means.
The cluster representation of the numerical values of the property data can also be seen as a symbolic qualitative measure representing each cluster.
Consequently, the number of clusters describes the granularity with which each property can qualitatively be represented. 
In case of a high number of clusters, an object is described in finer detail. Complementary, a lower number of clusters suggest a general description of an object. 
For instance, the numerical data about the \textit{\textbf{rigidity}} of the object instances of \textit{Ceramic Cup}, when clustered into three clusters, can be represented as $ \text{Rigidity} \mathrm{=} \{ \emph{soft}, \emph{medium}, \emph{rigid}\}$ (see Table~\ref{tab:discrete}).
At the end of the \textit{sub-categorization} process, each object is represented in terms of the qualitative measures for each property.

The conceptualization process gathers the knowledge about all the instances of an object class and represents the knowledge about an object class.
Initially, the knowledge about objects is represented using \textit{bivariate joint frequency distribution} of the qualitative measures of the properties in the object instances.
Next, conceptual knowledge about objects is calculated as a sample proportion of the frequency of the properties across the instances of a class.

The conceptual knowledge about instances and object classes is represented in JSON format in order to allow the users of the knowledge base to adapt the suitable representation formalism for their application.

\begin{table}
  \centering
  \caption{The sub-categorization process which generates the symbolic knowledge about object instances.}
  \label{tab:discrete}
  \begin{tabular}{|c|c||c|c|}
   \hline
   \multicolumn{2}{|c||}{Object} & Sensory Data & Discretized Data \\
   \hline
   Class & Instance & Rigidity & Rigidity \\
   \hline
  \multirow{3}{*}{Ceramic Cup} 	& ceramic\_cup\_1 & 0.76 & \emph{soft} \\ 
                        & ceramic\_cup\_2 & 3.17 & \emph{medium}\\
                        & ceramic\_cup\_3 & 7.69 & \emph{rigid}\\
   \hline
  \end{tabular}
\end{table}

\section{Preliminary Results}
\label{sec:preliminary}

In the endeavor of enabling a robot-centric conceptual knowledge acquisition, we introduced physical and functional properties of objects in the previous section and presented a framework implementing the envisioned process. 
Since this is a work in progress, we subsequently present preliminary results of the proposed dataset. %
For the initial experiments, we primarily focused on the evaluation of the discrimination of the object instances with regard to the acquired properties.
Given the acquired properties of each instance, we represented each instance in vector form, i.e. each vector element represents a specific physical property value.
For the preliminary results, we acquired the physical properties of 46 objects in total which span across 16 object classes. 
The number of instances considered for each class is stated in the Table~\ref{tab:dataset_label_definitions}.

\begin{figure}[tb]
	\centering
    \includegraphics[width=0.9\linewidth]{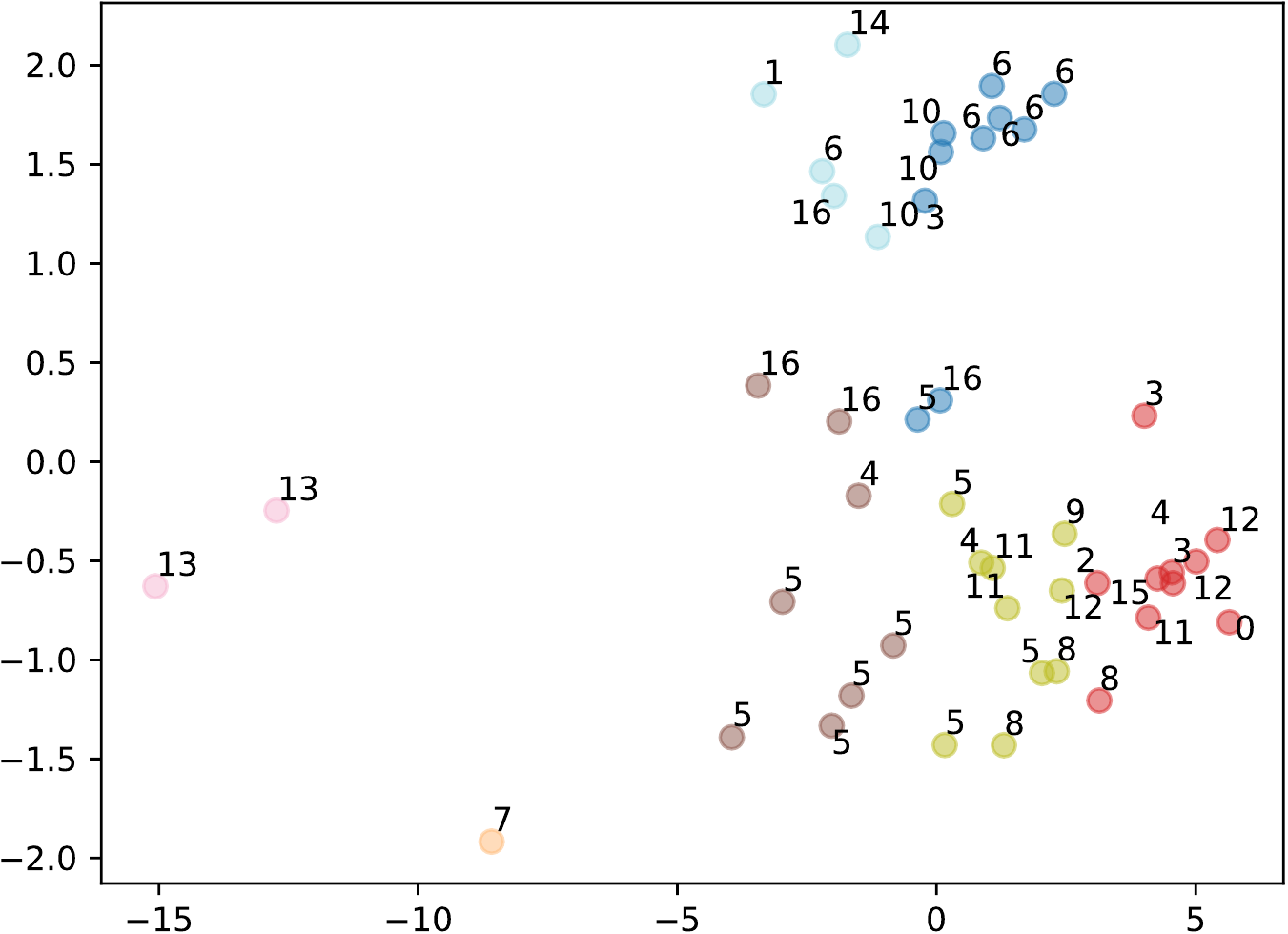}
	\caption{The instances are represented by the physical properties (reduced to two dimensional space~\cite{tenenbaum_global_2000}) and are clustered using K-means. Note: instances are colored w.r.t. to cluster assignment; cluster colors are randomly selected; instances are annotated with a numeric label according to Table~\ref{tab:dataset_label_definitions}.}
	\label{fig:p_isomap}
\end{figure}

The objective of this experiment is to examine whether the physical property measurements of the object instances convey the physical similarity between different objects.
For this experiment, each object instance was represented in terms of its physical properties.
The dimensionality of the instance data is initially reduced to two dimensions (see Fig.~\ref{fig:p_isomap}) using Isomap embedding technique~\cite{tenenbaum_global_2000}.
Next, the reduced data is split into seven clusters (half of the total number of actual object categories) using K-means clustering.
The results are depicted as a scatter plot in Fig.~\ref{fig:p_isomap}.
In the plot, an object instance is represented as a point which is colored according to its cluster assignment. Furthermore, each point is attributed with a numeric label according to its class label, see Table~\ref{tab:dataset_label_definitions}.
The clusters group together the instances which are physically similar, e.g., in the red cluster the instances of \textit{steel cup}, \textit{ceramic bowl}, \textit{plastic cup}, \textit{plastic box}, \textit{plastic bowl}, and \textit{ceramic cup} are physically similar according to the given set-up. 

\begin{figure}[tb]
	\centering
    \includegraphics[width=0.9\linewidth]{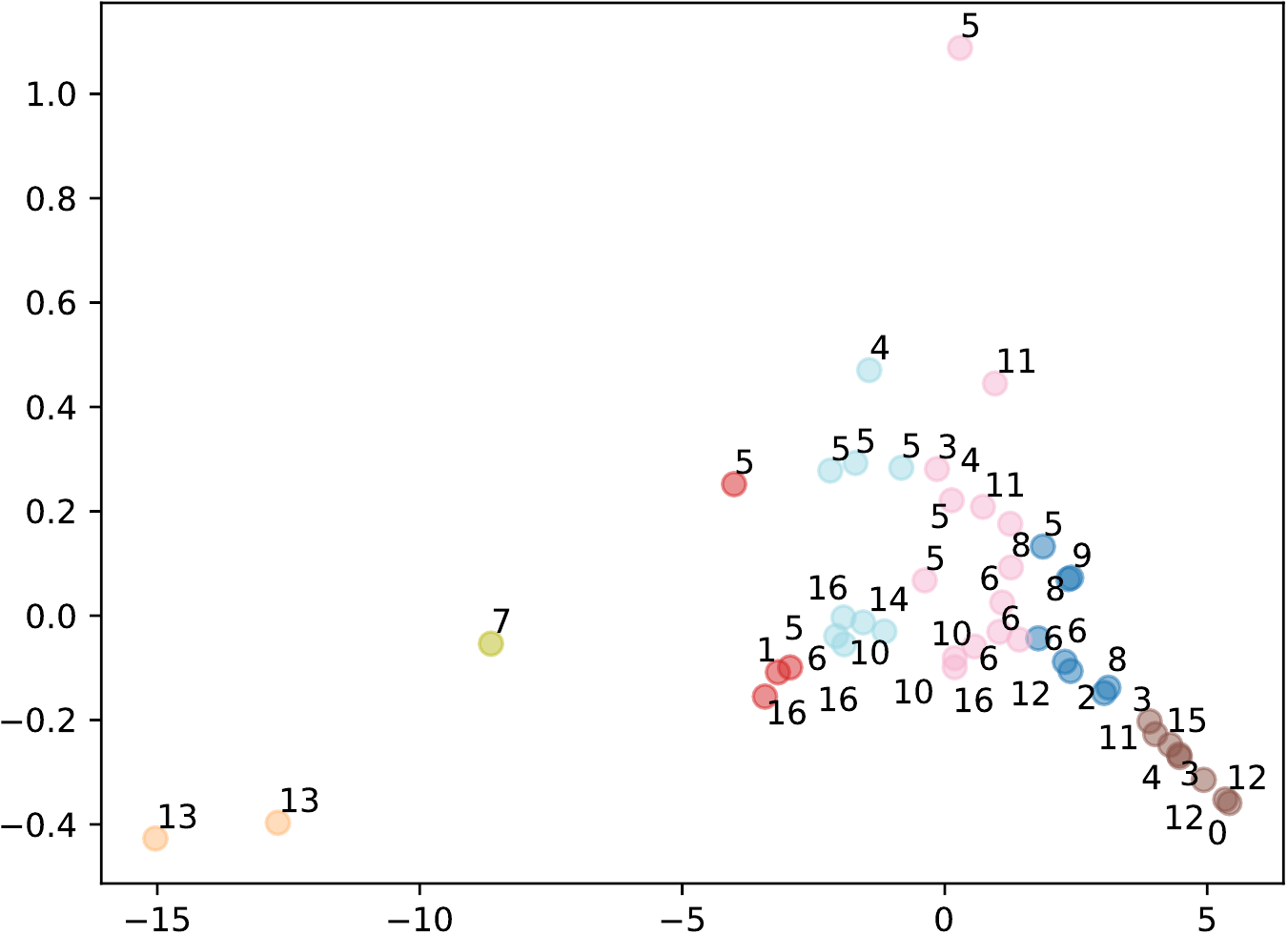}
	\caption{The instances are represented by support property (reduced to two dimensional space~\cite{tenenbaum_global_2000}) and are clustered using K-means. Note: instances are colored w.r.t. to cluster assignment; cluster colors are randomly selected; instances are annotated with a numeric label according to Table~\ref{tab:dataset_label_definitions}.}
	\label{fig:s_isomap}
\end{figure}

Similar experiments were conducted to evaluate the functional similarities between the objects.
Due to the lack of space, in Fig.~\ref{fig:s_isomap}, %
we have illustrated the similarity between objects with respect to \textit{\textbf{support}} functional property.
Accordingly, the instances of the object classes \textit{plastic bowl}, \textit{plastic cup}, \textit{ceramic bowl}, \textit{ceramic glass} and \textit{ceramic cup} represented by a brown cluster have the similar degree of \textit{\textbf{support}}.

\subsection{Limitations}
While designing the property extraction methods and the framework, we aimed for approaches that do not require sophisticated hardware and software components. Therefore, we use a marker-based instead of a generic object tracking in our experiments.
This mitigates the limitations that might be caused by complex approaches and facilitates robots to perform necessary experiments in an automated way. However, some limitations remain, which we discuss in this subsection.

One of the limitations is caused by mismatches between the definition of an object's property and the corresponding extraction method. For instance, instead of directly measuring the enclosed volume of an object to determine its \textit{\textbf{hollowness}}, we consider its internal depth and height.
Consequently, the extraction method will calculate the same value for an object with a spherical hollowness %
as it calculates for an object with a cylindrical hollowness of the same depth although the enclosed volume is different. A similar mismatch can occur when calculating the \textit{\textbf{size}} of an object, as we consider only its visual bounding box.

Another category of limitations are caused by the design of extraction methods. 
To measure the deformability of an object for extracting its \textit{\textbf{rigidity}}, it is placed on a planar, rigid surface. This surface itself provides \textbf{\textit{rigidity}} to objects (e.g. for a sheet of paper) and thereby causes incorrect measurements. Similarly, \textbf{\textit{roughness}} values for spherical objects are incorrect as these roll down the ramp instead of sliding.

Besides these methodological limitations, the employed hardware components impose limitations too. For instance, due to its size, the youBot's robotic arm does not allow extracting \textbf{\textit{rigidity}} values for objects with a width greater than 20\si{cm}. This ultimately limits the objects that can be analyzed.

\section{Current State and Future Direction}
\label{sec:future}
Standard datasets of the robotics community are generally created under supervision, one-dimensional, and discrete, i.e. an unary human-predefined label is given to an object sample; e.g. a  point cloud is labeled as a \emph{mug}.
The presented work focuses on a framework for generating a dataset from a robot-centric perspective by gathering continuous conceptual object knowledge such as the functional property \textit{\textbf{movability}}. 

We defined a set of object properties and their interrelations.
Therein we distinguish between physical and functional properties.
We show that these properties can be organized in a hierarchical bottom-up manner from low-level ones acquired from sensory data to high-level ones acquired from lower-level properties.
Given this basis, we proposed acquisition procedures for each 
property.
Eventually, we have introduced a framework consisting of property definitions and acquisition procedures and a corpus of 46 objects.

In our preliminary experiments we could show the discrimination of instances according to their physical and the functional property \textit{\textbf{support}}.
These observed results encourage us to continue on our goal of creating a robot-centric conceptual knowledge base. 

Therefore, we focus on extending the dataset with additional instances as well as classes in order to further investigate object understanding as such given the gained conceptual knowledge.
Furthermore we aim to mitigate the discussed limitations.
Considering the physical properties of \textit{\textbf{flatness}}, \textit{\textbf{hollowness}}, and \textit{\textbf{size}}, we plan for introducing 3D models of objects for generating more robust property values. 
Instead of directly processing noisy point clouds, a 3D model of an object will be created before extracting the respective properties.

While not considered in this early phase, failure modeling~\cite{Jaeger2018} and detection will be applied to enable failure-aware applications and to further mitigate the effects of sensor failures. Run-time approaches, such as the validity concept, were successfully applied to depth measurements of RGB-D sensors as well as to low-level distance sensors~\cite{Hoebel2017} and are specifically design to track sensor failures while propagating through a processing chain.

Finally, in the endeavor of enabling robots to perform the extraction methods autonomously, we plan for replacing the scale for measuring the objects weight. Instead, using the effort observations within the robotic arm can be used to determine the weight of an object while lifting it.

\bibliographystyle{IEEEtran}

\end{document}